\documentclass[conference]{IEEEtran}
\IEEEoverridecommandlockouts
\usepackage{cite}
\usepackage{amsmath,amssymb,amsfonts}
\usepackage{algorithmic}
\usepackage{graphicx}
\usepackage{textcomp}
\usepackage{xcolor}
\usepackage[ruled,vlined]{algorithm2e}
\usepackage{setspace}
\usepackage{multirow}
\usepackage{booktabs}
\usepackage{bigstrut}
\usepackage{hyperref}
\usepackage{xcolor}
\usepackage{mathrsfs}
\usepackage[english]{babel}
\usepackage{footnote}
\usepackage{enumitem}
\usepackage{hyperref}
\usepackage{graphicx}
\usepackage[caption=false]{subfig}
\usepackage{mathtools}
\usepackage{amsfonts}
\usepackage{amsthm}
\usepackage{diagbox}
\usepackage{flushend}

\newtheorem{definition}{\hspace{2em}Definition}

\newlength\savewidth\newcommand\shline{\noalign{\global\savewidth\arrayrulewidth
		\global\arrayrulewidth 1pt}\hline\noalign{\global\arrayrulewidth\savewidth}}

\usepackage{array}

\def\BibTeX{{\rm B\kern-.05em{\sc i\kern-.025em b}\kern-.08em
    T\kern-.1667em\lower.7ex\hbox{E}\kern-.125emX}}
\begin{document}

\title{CLDG: Contrastive Learning on Dynamic Graphs}

  


\author{
Yiming Xu\textsuperscript{1,2},
Bin Shi\textsuperscript{1,2}\IEEEauthorrefmark{1}\thanks{\IEEEauthorrefmark{1} Corresponding author.},
Teng Ma\textsuperscript{1,2},
Bo Dong\textsuperscript{2,3},
Haoyi Zhou\textsuperscript{4,5},
Qinghua Zheng\textsuperscript{1,2}\\
\textsuperscript{1}Department of Computer Science and Technology, Xi'an Jiaotong University, China\\
\textsuperscript{2} Shaanxi Provincial Key Laboratory of Big Data Knowledge Engineering, Xi'an Jiaotong University, China\\
\textsuperscript{3} Department of Distance Education, Xi'an Jiaotong University, China\\
\textsuperscript{4} School of Software, Beihang University, China\\
\textsuperscript{5} Advanced Innovation Center for Big Data and Brain Computing, Beihang University, China\\
\IEEEauthorblockA{\{xym0924, mateng0920\}@stu.xjtu.edu.cn, \{shibin, dong.bo, qhzheng\}@xjtu.edu.cn, haoyi@buaa.edu.cn}
}

\maketitle

\begin{abstract}
The graph with complex annotations is the most potent data type, whose constantly evolving motivates further exploration of the unsupervised dynamic graph representation. One of the representative paradigms is graph contrastive learning. It constructs self-supervised signals by maximizing the mutual information between the statistic graph's augmentation views. However, the semantics and labels may change within the augmentation process, causing a significant performance drop in downstream tasks. This drawback becomes greatly magnified on dynamic graphs. To address this problem, we designed a simple yet effective framework named CLDG. Firstly, we elaborate that dynamic graphs have temporal translation invariance at different levels. Then, we proposed a sampling layer to extract the temporally-persistent signals. It will encourage the node to maintain consistent local and global representations, i.e., temporal translation invariance under the timespan views. The extensive experiments demonstrate the effectiveness and efficiency of the method on seven datasets by outperforming eight unsupervised state-of-the-art baselines and showing competitiveness against four semi-supervised methods. Compared with the existing dynamic graph method, the number of model parameters and training time is reduced by an average of 2,001.86 times and 130.31 times on seven datasets, respectively. The code and data are available at: \url{https://github.com/yimingxu24/CLDG}.
\end{abstract}


\section{Introduction}
Graph neural network shows superior performance in many real-world applications, such as recommender systems~\cite{ying2018graph, wu2020graph, lu2020meta}, combinatorial optimization~\cite{li2018combinatorial, mirhoseini2020chip, cappart2021combinatorial}, trafﬁc prediction~\cite{yu2017spatio, guo2019attention, derrow2021eta}, risk management~\cite{hu2019cash, zhong2020financial, gao2021tax}, etc. Thus, it is widely investigated in recent years.
However, existing methods mainly focus on supervised or semi-supervised learning paradigms, which rely on abundant ground-truth label information. The acquisition of ground-truth labels on graphs may be more complex and difficult than other media forms. Because the graph is an abstraction of the real world, it is not as straightforward as video, image, or text. For example, even without considering data privacy and security in the transaction network, it is impossible to label whether the credit card is frauded through the crowdsourced manner, because it requires a wealth of expert knowledge and experience. These make it very expensive to obtain graph datasets with a large amount of label information. Therefore, it is significant and challenging work to learn rich representations on graphs with unsupervised methods.

Some recent works extend unsupervised contrastive learning to graphs~\cite{velickovic2019deep, hjelm2018learning, you2020graph, zhu2021graph, zhu2020deep, hassani2020contrastive, zhang2021canonical} to tackle this problem. They mainly generate augmented views by perturbing nodes, edges, and features, and then construct self-supervised signals by maximizing the mutual information (MI) between different augmented views, thereby eliminating the dependence on labels. 
However, the semantics and labels may change within the perturbation-based augmentation process, causing a significant performance drop in downstream tasks. For example, adding edges may introduce noise, and deleting edges may delete the most important edges and neighbors for nodes. 
More seriously, existing methods fail to utilize the temporal information when constructing comparison signals, which further limits the performance on dynamic graphs. 
Therefore, existing static graph contrastive learning methods may not be the optimal solution for the dynamic graph. How to generalize contrastive learning to dynamic graphs is a challenge.

\begin{figure}
  \centering
  \includegraphics[width=8.8cm]{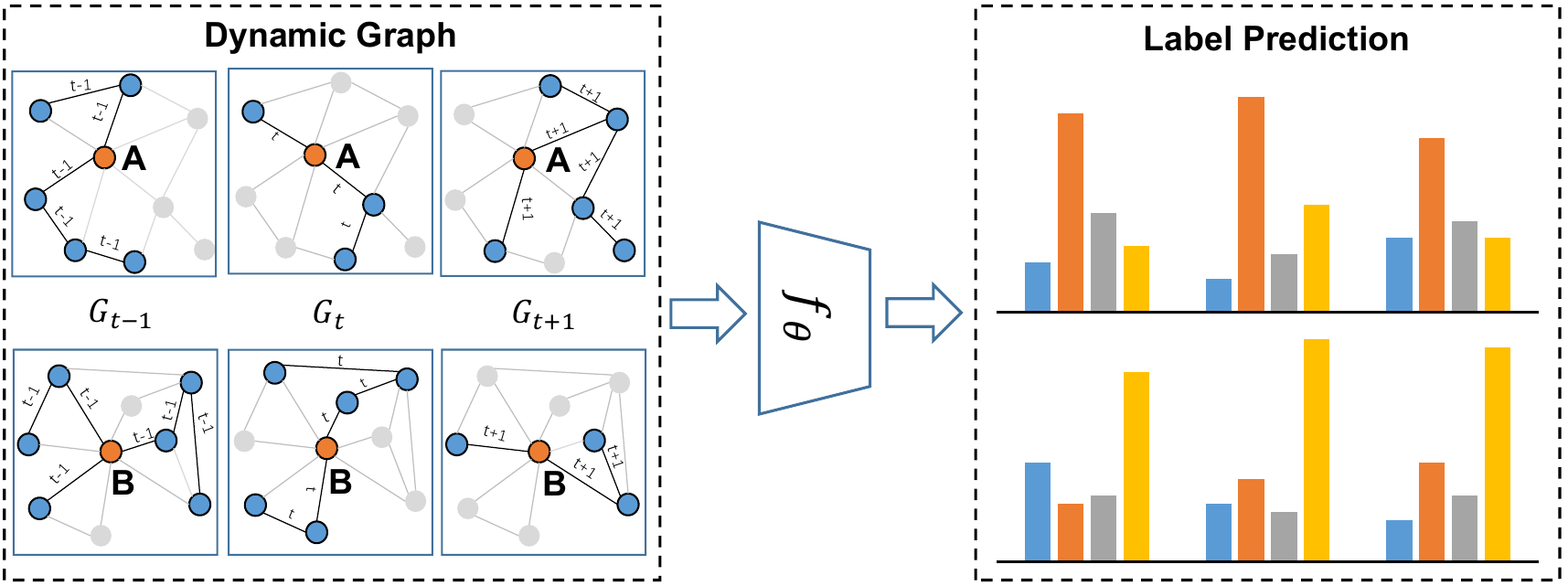}
  \caption{Illustration of our basic idea. In the dataset used in this paper (details of the datasets are demonstrated in Section~\ref{dataset}), an interesting observation is that the semantics and labels learned by the same node tend to be similar no matter what encoder is used at different timespans of the dynamic graph. We refer to this observation as temporal translation invariance. Based on temporal translation invariance, our basic idea is that the features of the node A, on different timespan should be similar and pull apart the features of other nodes, such as B. }
  \label{fig:TTI}
\vspace{-2mm}
\end{figure}

To achieve this objective, we perform some empirical studies to explore the properties of dynamic graphs (details of the experimental setup are demonstrated in Section~\ref{empirical}). An interesting observation is that the prediction labels of the same node tend to be similar in different timespans, regardless of the encoder used (GCN~\cite{kipf2016semi}, GAT~\cite{velivckovic2017graph} or MLP~\cite{mcculloch1943logical}), as shown in Figure~\ref{fig:TTI}.
The different datasets we use all share this characteristic (details of the datasets are demonstrated in Section~\ref{dataset}), and we refer to this observation on dynamic graphs as temporal translation invariance. Under the assumption of temporal translation invariance, we can utilize different timespan views to construct contrastive pairs. We implement two different levels (local and global) of temporal translation invariance. Specifically, at the local-level, we treat the semantics of the same node in different timespan views as positive pairs, encouraging the representation of the same node in each timespan view should be similar. At the global-level, the representation of a node and its neighbors in different timespan views is considered positive pairs. 
This approach is also intuitively reasonable. For example, in the academic network, most scientific researchers are deeply involved in a certain field and strive for it all their lives. Likewise, the business scope of a company does not change at will in the tax transaction network. 
In this case, different timespan views of the dynamic graph can be regarded as a more graceful contrastive pairs selection method, without manual trial-and-error, cumbersome search, or expensive domain knowledge for augmentation selection~\cite{xia2022simgrace}. Meanwhile, the additional temporal cues provided by dynamic graphs are implicitly exploited by maintaining temporal translation invariance on the graph.

To sum up, we propose a new inductive bias on dynamic graphs, namely temporal translation invariance. We design a conceptually simple yet effective unsupervised \textbf{C}ontrastive \textbf{L}earning framework on the \textbf{D}ynamic \textbf{G}raph, called CLDG. Specifically, we first generate multiple views from continuous dynamic graphs via a timespan view sampling method. Subsequently, feature representations of nodes and neighborhoods are learned through a weight-shared encoder and readout function and a weight-shared projection head. Finally, we design local-level and global-level contrastive losses separately to train the model. The pseudocode for CLDG is provided in Algorithm~\ref{alg:CLDG_algo}. 

The main contributions are summarized as follows: 

$\bullet$ We propose an \textbf{efficient} contrastive learning method on dynamic graphs to perform representation learning on discrete-time and continuous-time dynamic graphs in unsupervised scenarios.

$\bullet$ The proposed method is \textbf{simpler} and \textbf{lighter}, and has lower space and time complexity than existing dynamic graph methods.

$\bullet$ The proposed method is \textbf{highly scalable}, allowing the selection of various encoder architectures in the past, present and even in the future without any restrictions.

$\bullet$ The experimental results on 7 datasets demonstrated that CLDG yields state-of-the-art results of unsupervised dynamic graph representation learning on 12 established baselines, and even achieves better classification accuracy than supervised learning methods on 4 datasets.

\definecolor{commentcolor}{RGB}{46,139,87}   
\newcommand{\PyComment}[1]{\ttfamily\textcolor{commentcolor}{\# #1}}  
\newcommand{\PyCode}[1]{\ttfamily\textcolor{black}{#1}} 
\begin{algorithm}[!htb]\small
\setstretch{1.08}
\SetAlgoLined
    \PyComment{f: encoder networks + projection head} \\
    \PyComment{g: input graph} \\
    \PyComment{N: batchsize} \\ 
    \PyComment{d: embedding dimension} \\
    \PyComment{t: temperature} \\
    \leavevmode \\
    \PyComment{Timespan View Sampling Layer} \\
    \PyCode{g1, feat1 = temporal\_sampling(g)} \\ 
    \PyCode{g2, feat2 = temporal\_sampling(g)} \\ 
    \PyCode{z1 = f(g1, feat1)} \PyComment{N×d} \\ 
    \PyCode{z2 = f(g2, feat2)} \PyComment{N×d} \\ 
    
    \leavevmode \\
    \PyComment{\footnotesize temporal translation invariance} \\
    \PyCode{loss = (ctr(z1,z2) + ctr(z2, z1)) / 2} \\
    \PyCode{loss.backward()} \\
    \leavevmode \\
    \PyComment{contrastive loss} \\
    \PyCode{def ctr(q, k):} \\
    \Indp   
        \PyCode{logits = mm(q, k.T)} \PyComment{N×N} \\ 
        \PyCode{labels = range(N)} \PyComment{positives are in diagonal} \\ 
        \PyCode{loss = CrossEntropyLoss(logits / t, labels)} \\
        \PyCode{return loss} \\
    \Indm 
\caption{Pseudocode of CLDG in a PyTorch-like style.}
\label{alg:CLDG_algo}
\end{algorithm}

\section{Related Work}
\subsection{Contrastive Learning on Graphs}
Supervised and semi-supervised learning still dominate in graph neural networks~\cite{verma2019graphmix, feng2020graph, wang2020graph}. However, the graph itself is an abstraction of the real world, and graph data suffers from the problems of lack of labels and difficulty in labeling. Recently, contrastive learning is highly successful in computer vision (CV) and natural language processing (NLP)~\cite{wu2018unsupervised, van2018representation, he2020momentum, chen2020simple, grill2020bootstrap, gao2021simcse}. Inspired by the above methods, there are some works that extend contrastive learning to graphs. DGI~\cite{velickovic2019deep} extends deep InfoMax~\cite{hjelm2018learning} to graphs and maximizes MI between global graph embeddings and local node embeddings. GraphCL~\cite{you2020graph} proposes a novel graph contrastive learning framework that systematically explores the performance impact of various combinations of four different data augmentations. GCA~\cite{zhu2021graph} and GRACE~\cite{zhu2020deep} focus more on the graph data augmentation, and propose an adaptive data augmentation scheme at both topology and node attribute. MVGRL~\cite{hassani2020contrastive} creates another view for the graph by introducing graph diffusion~\cite{klicpera2019diffusion}. A discriminator contrasts node representations from one view with graph representation of another view and vice versa, and scores the agreement between representations which is used as the training signal. CCA-SSG~\cite{zhang2021canonical} first generates two views of the input graph through data augmentation, and then uses the idea based on Canonical Correlation Analysis (CCA) to maximize the correlation between the two views and encourages different feature dimensions to capture distinct semantics. However, the above methods have two problems: (1) They require two views generated by corrupting the original graph, such as node perturbation, edge perturbation and feature perturbation, etc. Inappropriate data augmentation may introduce noisy information resulting in semantic and label changes. (2) They are designed for static graphs. Applying contrastive learning directly to dynamic graphs is not straightforward. 

\subsection{Representation Learning on Dynamic Graphs}
Traditionally, the static graph representation problem is intensively studied by researchers and a variety of effective works are proposed~\cite{kipf2016semi, velivckovic2017graph, tang2015line, perozzi2014deepwalk, huang2020combining, chien2021node}. However, real-world networks all evolve over time, which poses important challenges for learning and inference. Therefore, there is an increasing amount of research on dynamic graph representations recently. According to the modeling method of the dynamic graph, these works can be roughly divided into discrete-time methods~\cite{goyal2018dyngem, sankar2020dysat, xu2020inductive, pareja2020evolvegcn, xue2020modeling, wang2021temporal} and continuous-time methods~\cite{zuo2018embedding, kumar2019predicting, trivedi2019dyrep, lu2019temporal}.

\noindent\textbf{Discrete-time Methods} 
DynGEM~\cite{goyal2018dyngem} uses a deep autoencoder to capture the connectivity trends in a graph snapshot at any time step. DySAT~\cite{sankar2020dysat} generates dynamic node representations through self-attention along both structural and temporal. TGAT~\cite{xu2020inductive} uses the self-attention mechanism and proposes a time encoding technique based on the theorem of Bochner. STAR~\cite{xu2019spatio}, DyHATR~\cite{xue2020modeling}, dyngraph2vec~\cite{goyal2020dyngraph2vec} and TemGNN~\cite{wang2021temporal} use different encoders (such as GCN~\cite{kipf2016semi}, autoencoder~\cite{baldi2012autoencoders}, hierarchical attention model, etc.) to extract the features of each time snapshot respectively, and then introduce sequence models such as LSTM~\cite{hochreiter1997long} and GRU~\cite{cho2014learning} to capture timing information. 

\noindent\textbf{Continuous-time Methods} 
HTNE~\cite{zuo2018embedding} proposes a Hawkes process based temporal network embedding method to capture the influence of historical neighbors on the current neighbors. JODIE~\cite{kumar2019predicting} uses a coupled RNN architecture to update the embedding of users and items at every interaction. DyRep~\cite{trivedi2019dyrep} builds a two-time scale deep temporal point process approach to capture the continuous-time fine-grained temporal dynamics processes. M$^2$DNE~\cite{lu2019temporal} designs a temporal attention point process to capture the fine-grained structural and temporal properties in micro-dynamics, and defines a dynamics equation to impose constraints on the network embedding in macro-dynamics.

However, existing dynamic graph work still suffers from at least one of the following limitations: (1) Most RNN-like methods have high time complexity and are not easy to parallel. This hinders the scaling of existing dynamic graph models to large-scale graphs. (2) Most models by reconstructing future states or temporal point processes or sequences may learn noisy information as they try to fit each new interaction in turn. (3) The temporal regularizer is similar to contrastive learning without negative examples, which forces the node representation to smooth from adjacent snapshots. However, a potential problem of this approach is the existence of completely collapsed solutions.

\section{Preliminaries}
In this section, we give the necessary notations and definitions used throughout this paper. The main notations are summarized in Table~\ref{tab:ND}.

\subsection{Dynamic Graph Modeling}
The existing dynamic graph modeling methods can be roughly divided into two categories~\cite{xue2022dynamic, kazemi2020representation}: discrete-time dynamic graph and continuous-time dynamic graph. We formally define two modeling methods as follows:

\begin{definition} [\textbf{Discrete-time Dynamic Graph}]
A discrete-time dynamic graph (DTDG) is a sequence of network snapshots within a given time interval. Formally, we define a DTDG as a set $\left \{\mathcal{G}^{1}, \mathcal{G }^{2},... ,\mathcal{G }^{T} \right \}$ where $\mathcal{G}^{t}=\left \{\mathcal{V}^{t}, \mathcal{E}^{t}\right \}$ is the graph at snapshot $t$, $\mathcal{V}^t$ is the set of nodes in $\mathcal{G}^{t}$, and $\mathcal{E}^t\subseteq \mathcal{V}^t\times \mathcal{V}^t$ is the set of edges in $\mathcal{G}^{t}$.
\end{definition}

\begin{definition} [\textbf{Continuous-time Dynamic Graph}]
A continuous-time dynamic graph (CTDG) is a network with edges and nodes annotated with timestamps. Formally, we define a CTDG as $\mathcal{G}=\left \{\mathcal{V}^T, \mathcal{E}^T, \mathcal{T}\right \}$ where $\mathcal{T}:\mathcal{V},\mathcal{E}\rightarrow \mathbb{R}^{+}$ is a function that maps each node and edge to a corresponding timestamp.
\end{definition}

\subsection{Problem Formulation}
The objective of this paper is to design a dynamic graph representation (embedding) method. The mathematical formulation of this problem could be defined as:

\emph{Given a dynamic graph $\mathcal{G}=\left \{\mathcal{V}, \mathcal{E}, \mathcal{T}\right \}$, our goal is to learn a mapping function $f:v_{i}\to \mathbf{z}_{i}\in \mathbb{R}^{d}$, where $d\ll \left | \mathcal{V}\right |$, and $\mathbf{z}_{i}$ is the embedded vector that preserves both temporal and structural information of vertex $v_{i}$.}

\newcommand{\tabincell}[2]{\begin{tabular}{@{}#1@{}}#2\end{tabular}} 
\begin{table}\small
\centering
\renewcommand\arraystretch{1.2}
  \caption{Notations and Descriptions.}
  \label{tab:ND}
  \begin{tabular}{cc}
    \toprule
    Notations & Descriptions\\
    \midrule
    $\mathcal{G}$ & A dynamic graph \\
    $\mathcal{V}$ & The set of nodes in $\mathcal{G}$ \\
    $\mathcal{E}$ & The set of edges in $\mathcal{G}$ \\
    $\tau$  & The temperature parameter \\
    $s$ & The view timespan factor \\
    $v$ & The number of views sampled \\
    $a,\mathbf{a},\mathbf{A}$ & Scalar, vector, matrix\\
    $\mathcal{N}_{i}^{\mathcal{T}}$  & The neighbor set of node $i$ under view $\mathcal{T}$\\
    $\mathbb{R}_{\textrm{sample}}\left ( \cdot ,\cdot ,\cdot  \right )$ & Timespan view sampling function \\
  \bottomrule
\end{tabular}
\end{table}

\begin{figure*}
  \centering
  \includegraphics[width=16cm]{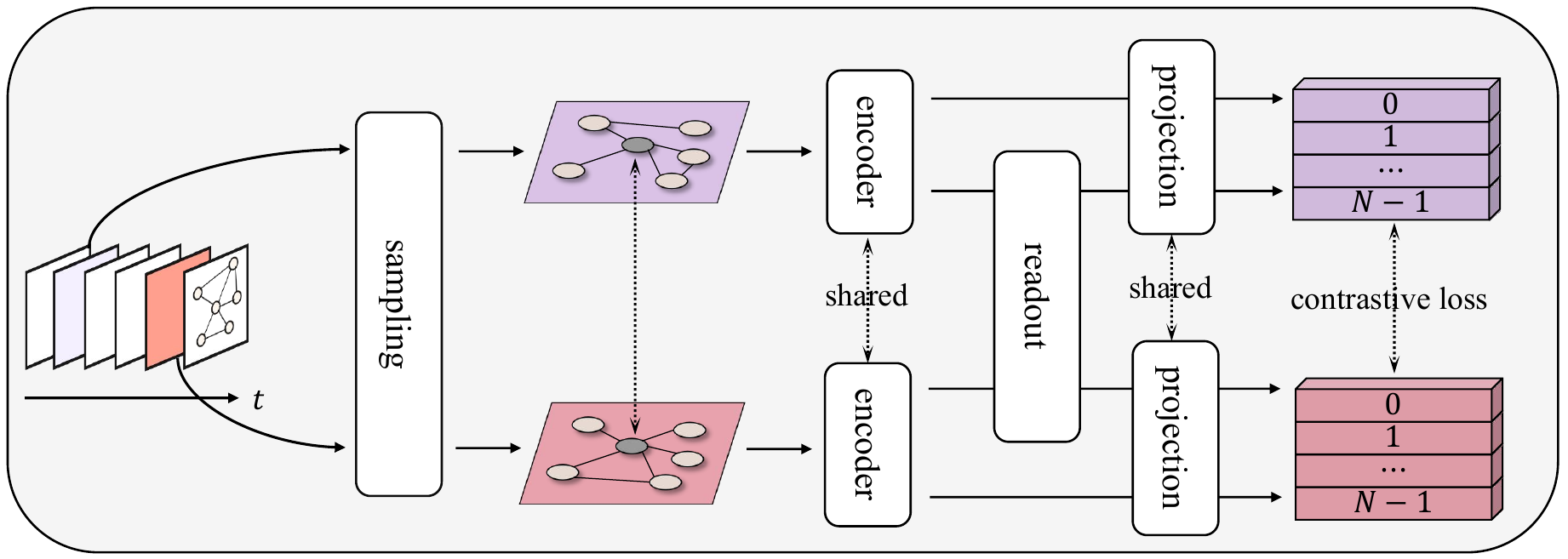}
  \caption{The architecture of the CLDG. The core of CLDG is implemented by maintaining local or global temporal translation invariance. Given an input graph, we first sample multiple views through a timespan view sampling layer. Subsequently, the multi-views are fed into a shared weight encoder to generate node embeddings, and the node neighborhood embedding is generated through the readout function. Then, node and neighborhood embeddings are fed into a projection head that maps the embeddings into the space of the contrastive loss. Finally, the contrastive approach is used to maintain the temporal translation invariance of the local and the global with a batch size of $N$.}
  \label{fig:overview}
\end{figure*}

\section{Proposed Method}
In this section, we propose a new unsupervised dynamic graph representation learning method that addresses two progressive challenges: (1) how to apply contrastive learning to dynamic graphs (presented in IV.A); (2) based on the solution of the first challenge, how to specifically select timespans as contrastive pairs (presented in IV.B).

The framework overview and workflow of CLDG are shown in Figure~\ref{fig:overview}. It consists of five major lightweight components: timespan view sampling layer, base encoder, readout function, projection head and contrastive loss function. To begin with, the view sampling layer extracts the temporally-persistent signals. Then, the base encoder and the readout function learn the local and global representations. Furthermore, the projection head maps the representations into the space of the contrastive loss. Finally, the contrastive loss function is used to maintain the temporal translation invariance of the local and the global.

\subsection{Temporal Translation Invariance}\label{empirical}
CLDG aims to generalize contrastive learning to dynamic graphs, where the key challenge is to select proper contrastive pairs. 
In the field of NLP and CV, the common solution for this issue is to generate different views of the samples as positive pairs through data augmentation. It shows good performance since the augmentation methods generate semantically invariant positive pairs. For example, a leopard in an image remains a leopard after geometric transformations, such as flip, rotate, crop, scale, pan, dither, etc., and pixel transformation methods, such as noise, adjust contrast, brightness, saturation, etc. However, the graph is an abstraction of the real world, and existing graph data augmentation methods such as node perturbation, edge perturbation, and feature perturbation may cause the labels of nodes or graphs to change. For example, with edge perturbation in a citation network, adding edges may cause completely irrelevant papers to be linked together, and edge dropping may drop edges and neighbors that are most important to a node. Therefore, proposing a more graceful contrastive pairs selection method is of great importance in our problem.

In fact, there are plenty of additional cues provided by dynamic graphs in the temporal component. To explore the properties of dynamic graphs, we perform some empirical studies. Specifically, we first process seven dynamic graph datasets $\mathcal{G}$ by spliting each of them into many timespans $\left \{\mathcal{G}^{1}, \mathcal{G }^{2},... ,\mathcal{G }^{T} \right \}$ (the specific datasets described in Section~\ref{dataset}). For each timespans, we train common GNN encoders with the same architecture but without shared weights. The trained model is then used to predict the node labels. An interesting observation is that regardless of the encoder used for training, the prediction labels of the same node tend to be similar in different timespans, as shown in Figure~\ref{fig:TTI}. We refer to this phenomenon as temporal translation invariance.

Under the assumption of temporal translation invariance, we can utilize different timespan views to construct local (node pairs) and global (node-graph pairs) contrastive pairs. Specifically, we implement local-level and global-level temporal translation invariance. In local-level temporal translation invariance, we treat the semantics of the same node in different timespan views as positive pairs, pulling closer the representation of the same node in each timespan view, and pulling apart the representation of different nodes. In global-level temporal translation invariance, we treat the semantics of the node and its neighbor in different timespan views as a positive pair, pulling closer the representation of a node and its neighbor at different timespan views, and pulling apart the neighbor representation of other nodes.

\subsection{Timespan View Sampling Layer}\label{tpgs}
\begin{figure}
  \centering
  \includegraphics[width=9.0cm]{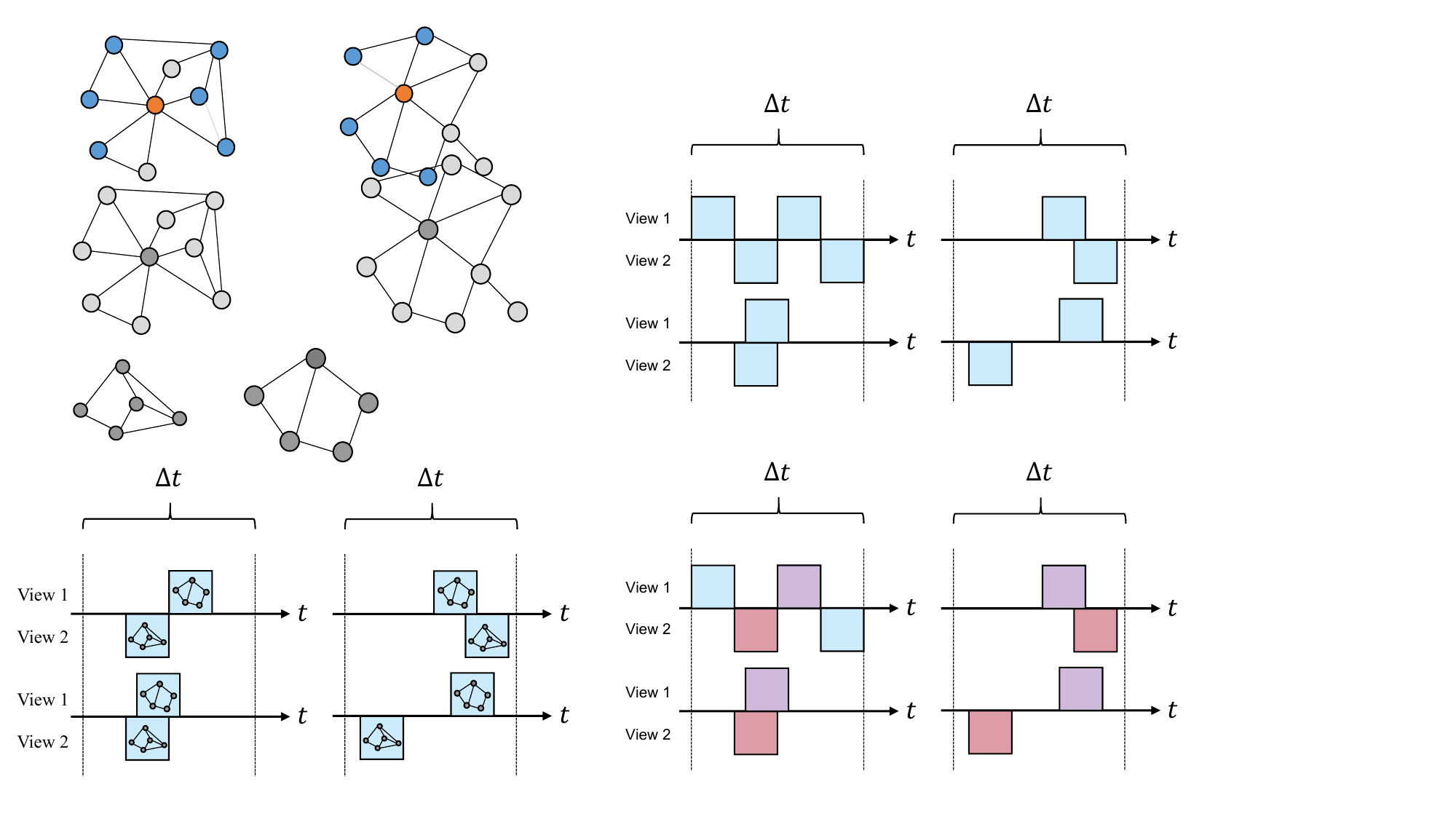}
  \caption{Four candidate timespan view sampling strategies of CLDG.}
  \label{fig:sampling}
\end{figure}

Based on temporal translation invariance, CLDG utilizes different timespan views as contrastive pairs. However, how to specifically choose appropriate timespans is still an open question.
First, it is intuitive that the interval distance between different timespan views may affect the contrastive learning results. In the case of two timespan views, if more temporal overlap exists between two views, they theoretically share more similar semantic contexts, but this may lead to over-simplified tasks. If two views are separated for a long time, they may have completely different neighbors and may no longer share the same semantic context. In addition, how to choose the appropriate number and size of timespan views is also to be studied. Therefore, we design four different timespan view sampling strategies to explore the optimal view interval distance selection, as shown in Figure~\ref{fig:sampling}. The main difference between each strategy is the rate of physical temporal overlap, thereby sharing different semantic contexts. Meanwhile, each sampling strategy considers both the number and size of views. Next, we formally define the timespan view sampling layer.

Specifically, given a continuous-time dynamic graph $\mathcal{G}$, we first define that the overall timespan of graph $\mathcal{G}$ is $\Delta t=\max{\left ( \mathcal{T} \right )}-\min{\left ( \mathcal{T} \right )}$. Then, we set two factors $v,s$ to control the number and size of the sampling timespan views respectively, i.e., $v$ views are sampled for each strategy and the timespan of each view is $\Delta t / s$, and $v,s\in \mathbb{N^{*}}$ are hyperparameters that can be set according to the actual physical implication of the graph and the density of the graph. Finally, the timespan view sampling strategy can be formulated as:
\begin{equation}
\left ( \mathcal{T}_{1},\cdots ,\mathcal{T}_{v} \right )=\mathbb{R}_{\textrm{sample} }\left (s ,v,\Delta t \right ),
\end{equation}
where $\mathbb{R}_{\textrm{sample} }\left ( \cdot ,\cdot ,\cdot  \right )$ returns a sample time tuple, where $\mathcal{T}_{i}\in \left [ \min{\left ( \mathcal{T} \right )}+\frac{\Delta t}{2s},\max{\left ( \mathcal{T} \right )}-\frac{\Delta t}{2s} \right ], \forall i\in \left [ 1,v \right ]$. 

We obtain $v$ timespan views, i.e., $\widetilde{\mathcal{G}}_{1}, \widetilde{\mathcal{G}}_{2},\cdots, \widetilde{\mathcal{G}}_{v}$, according to the sample time tuples $\left ( \mathcal{T}_{1},\cdots ,\mathcal{T}_{v} \right )$, where $\widetilde{\mathcal{G}}_{i}$ retains all edges of $\mathcal{G}$ in  a specific timeframe $\left [ \mathcal{T}_{i}-\frac{\Delta t}{2s},\mathcal{T}_{i}+\frac{\Delta t}{2s} \right ]$.
Specifically, the sample time tuple can be generated by four different strategies. 
We formally define four sampling strategies by the time difference between $\mathcal{T}_{i}$ in the sample time tuple and its predecessor $\mathcal{T}_{i-1}$ and successor $\mathcal{T}_{i+1}$. 

\textbf{Sequential Sampling Strategy}. 
This strategy first divides the dynamic graph into $s$ timespan views, with no intersection between each view, and then randomly samples $v$ unduplicated views, where $v\leq s$. This strategy is similar to the processing method of DTDG, which can be directly used in DTDG. 
\begin{equation}
\left| \mathcal{T}_{i}-\mathcal{T}_{i\pm 1} \right| = \alpha \frac{\Delta t}{s},
\end{equation}
where $\alpha \in \mathbb{N^{*}}$, the time difference between any two views is an integer multiple of the timespan of the view. 

\textbf{High Overlap Rate Sampling Strategy}.
The $v$ views sampled by this strategy are interdependent. We set an overlap rate of 75\%, that is, the time between the sampled dynamic graphs $\widetilde{\mathcal{G}}_{i}$ and $\widetilde{\mathcal{G}}_{i\pm1}$ corresponding to $\mathcal{T}_{i}$ and $\mathcal{T}_{i\pm 1}$ has a 75\% overlap. 
\begin{equation}
\left| \mathcal{T}_{i}-\mathcal{T}_{i\pm 1} \right| = \frac{\Delta t}{4s},
\label{eq:high_overlap}
\end{equation}
where $\mathcal{T}_{1}$ is limited by $\left [ \min{\left ( \mathcal{T} \right )}+\frac{\Delta t}{2s},\max{\left ( \mathcal{T} \right )}-\frac{\left ( 2+v \right )\cdot \Delta t}{4s} \right ]$ and random sampling in each epoch, assuming that $\mathcal{T}_{1}< \mathcal{T}_{i} < \mathcal{T}_{v}$ is satisfied in the sampled time tuple. Then $\mathcal{T}_{2}$ to $\mathcal{T}_{v}$ are obtained according to Eq.~\ref{eq:high_overlap}.

\textbf{Low Overlap Rate Sampling Strategy}.
The $v$ views sampled by this strategy are interdependent. We set an overlap rate of 25\%, which means that the time between the sampled dynamic graphs $\widetilde{\mathcal{G}}_{i}$ and $\widetilde{\mathcal{G}}_{i\pm1}$ corresponding to $\mathcal{T}_{i}$ and $\mathcal{T}_{i\pm 1}$ has a 25\% overlap. 
\begin{equation}
\left| \mathcal{T}_{i}-\mathcal{T}_{i\pm 1} \right| = \frac{3\Delta t}{4s},
\label{eq:low_overlap}
\end{equation}
where $\mathcal{T}_{1}\in \left [ \min{\left ( \mathcal{T} \right )}+\frac{\Delta t}{2s},\max{\left ( \mathcal{T} \right )}-\frac{\left ( 2+3v \right )\cdot \Delta t}{4s} \right ]$ and random sampling in each epoch, assuming that $\mathcal{T}_{1}< \mathcal{T}_{i} < \mathcal{T}_{v}$ is satisfied in the sampled time tuple. Then $\mathcal{T}_{2}$ to $\mathcal{T}_{v}$ are obtained according to Eq.~\ref{eq:low_overlap}.

\textbf{Random Sampling Strategy}.
$\mathbb{R}_{\textrm{sample} }\left ( \cdot ,\cdot ,\cdot  \right )$ randomly returns a set of sample time tuples, where $\forall i $ satisfies $\mathcal{T}_{i}\in \left [ \min{\left ( \mathcal{T} \right )}+\frac{\Delta t}{2s},\max{\left ( \mathcal{T} \right )}-\frac{\Delta t}{2s} \right ]$. The dynamic graphs $\widetilde{\mathcal{G}}_{i}$ and $\widetilde{\mathcal{G}}_{j}$ sampled corresponding to $\mathcal{T}_{i}$ and $\mathcal{T}_{j}$ may not overlap or partially overlap in time.
\begin{equation}
\left| \mathcal{T}_{i}-\mathcal{T}_{i\pm 1} \right| \in \left [ 0,\Delta t - \frac{\Delta t}{s} \right ].
\end{equation}

\subsection{Base Encoder Layer}
In each training epoch, the timespan view sampling layer first samples $V$ graph views, defined as $\left\{\widetilde{\mathcal{G}}_{1},\widetilde{\mathcal{G}}_{2},\cdots ,\widetilde{\mathcal{G}}_{V} \right\}$ and $\widetilde{\mathcal{G}}_{i}=\left\{\mathbf{X}_{i},\mathbf{A}_{i}\right\}$, where $\mathbf{X}_{i}$ and $\mathbf{A}_{i}$ are the feature matrix and adjacency matrix of the i-th view. For each view, CLDG allows arbitrary encoder network architecture selection without any restrictions. We can even use the encoder of existing dynamic graphs. In this paper, for simplicity, we choose GCN~\cite{kipf2016semi} as the base encoder to model the structural dependencies. The multi-layer graph convolution of GCN is defined as follows: 

\begin{equation}
\mathbf{H}^{\left ( l+1 \right )}=\sigma \left (\widetilde{\mathbf{D}}^{-\frac{1}{2}}\widetilde{\mathbf{A}}\widetilde{\mathbf{D}}^{-\frac{1}{2}}\mathbf{H}^{\left ( l \right )}\mathbf{W}^{\left ( l \right )} \right ),
\label{eq:encoder}
\end{equation}
where for each view $\widetilde{\mathcal{G}}_{i}$, $\widetilde{\mathbf{A}}=\mathbf{A}_{i}+\mathbf{I}_{N}$ and $\mathbf{I}_{N}$ is the identity matrix, $\widetilde{\mathbf{D}}$ is the degree matrix of $\widetilde{\mathbf{A}}$, $\mathbf{H}^{\left ( 0 \right )}= \mathbf{X}_{i}$. $\mathbf{W}^{\left ( l \right )}$ is a layer-specific trainable transformation matrix. It is worth noting that the different views sampled on the dynamic graph according to the timespan view sampling layer contain different sets of nodes and edges. Therefore, we used the minibatch training approach in the specific implementation. Simultaneously, to ensure better consistency of the learned representations among views, we input all views into the encoder with shared weights, and the node embedding output by the i-th view is $\mathbf{H}_{i} = f_{\theta } \left ( \mathbf{X}_{i},\mathbf{A}_{i} \right )$, where $\mathbf{H}_{i} \in \mathbb{R}^{N \times d}$, $N$ is the batch size, $d$ is the embedding dimension. 

\subsection{Readout Function Layer}
After the base encoder layer, we get the embedding of each node. To maintain the global temporal translation invariance on the graph, i.e., a node and its neighbors at different views constitute a positive pair, it is necessary to consider how to generate the neighbor representations of nodes. Some works achieve state-of-the-art on graph classification tasks by designing more efficient readout functions through differentiable pooling~\cite{ying2018hierarchical} or a learnable filter function~\cite{hofer2020graph}. However, the graph pooling operation is not the core of this work. Therefore, we still use statistical-based methods such as average, maximum, and summation. This approach is simple and effective, while not introducing additional model parameters. The formal expression is as follows:
\begin{equation}
\mathbf{h}_{\mathcal{N}_{i}}^{\mathcal{T}}= \textrm{readout} \left ( \left\{ \mathbf{h}_{j}^{\mathcal{T}},\forall j\in \mathcal{N}_{i}^{\mathcal{T}}\right\} \right ),
\end{equation}
where $\mathcal{N}_{i}^{\mathcal{T}}$ is the neighbor set of node $i$ under view $\mathcal{T}$. $\mathbf{h}_{\mathcal{N}_{i}}^{\mathcal{T}}$ is the neighbor embedding representation of node $i$ under $\mathcal{T}$.

\begin{figure}
  \centering
  \includegraphics[width=7.5cm]{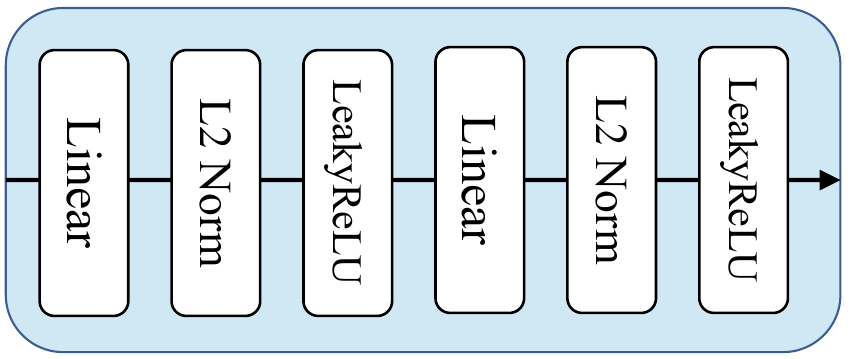}
  \caption{Projection head of CLDG.}
  \label{fig:projection}
\end{figure}

\subsection{Projection Head Layer}
The base encoder and readout function extract the representation of nodes and node neighborhoods from the sampled time views. Then, we feed the representations into a shared weight projection head, which maps the representation to the space where contrastive loss is applied. The projection head formula is as follows: 
\begin{equation}
\mathbf{z}_{i}^{\mathcal{T}},\mathbf{z}_{\mathcal{N}_{i}}^{\mathcal{T}_{k}}=\textrm{proj} \left (  \mathbf{h}_{i}^{\mathcal{T}} \right ),\textrm{proj} \left (  \mathbf{h}_{\mathcal{N}_{i}}^{\mathcal{T}} \right ) ,
\end{equation}
where $\textrm{proj} \left ( \cdot  \right ) $ is an MLP with two layers, $l_{2}$ normalized and LeakyReLU non-linearity, as shown in Figure~\ref{fig:projection}.

\subsection{Contrastive Loss Function}
Based on the observation that node labels tend to be invariant over the entire dynamic graph time period, we propose a new inductive bias on dynamic graphs, i.e., local and global temporal translation invariance. Distinguishing from previous work, we learn node embedding by maximizing the temporal consistency between local or global. 
Specifically, in local temporal translation invariance, we treat the semantics of the same node in different timespan views as positive pairs,  and different nodes as negative pairs. In global temporal translation invariance, we treat the semantics of the node and its neighbor in different timespan views as a positive pair, and the neighbor of other nodes as negative pairs.
In this paper, we use the InfoNCE~\cite{van2018representation} contrastive approach. The objective of local temporal translation invariance is defined as follows: 

\begin{equation}
\mathcal{L}_{i}^{\textrm{node} }=-\log\frac{\exp{\left ( \mathbf{z}_{i}^{\mathcal{T}_{q}}\cdot \mathbf{z}_{i}^{\mathcal{T}_{k}}/\tau  \right )}}{\sum_{j=1}^{N}\exp{\left ( \mathbf{z}_{i}^{\mathcal{T}_{q}}\cdot \mathbf{z}_{j}^{\mathcal{T}_{k}}/\tau  \right )}},
\end{equation}
where $\mathbf{z}_{i}^{\mathcal{T}_{q}}$ and $\mathbf{z}_{i}^{\mathcal{T}_{k}}$ are the representations learned by node $i$ in views $\mathcal{T}_{q}$ and $\mathcal{T}_{k}$ through the encoder and projection head, and $q\neq k$. $N$ is the batch size, and $\tau$ is a temperature parameter. 

The objective of global temporal translation invariance is defined as follows: 

\begin{equation}
\mathcal{L}_{i}^{\textrm{graph} }=-\log\frac{\exp{\left ( \mathbf{z}_{i}^{\mathcal{T}_{q}}\cdot \mathbf{z}_{\mathcal{N}_{i}}^{\mathcal{T}_{k}}/\tau  \right )}}{\sum_{j=1}^{N}\exp{\left ( \mathbf{z}_{i}^{\mathcal{T}_{q}}\cdot \mathbf{z}_{\mathcal{N}_{j}}^{\mathcal{T}_{k}}/\tau  \right ) }},
\end{equation}
where $\mathbf{z}_{\mathcal{N}_{i}}^{\mathcal{T}_{k}}$ is the representation of the neighbors of node $i$ in snapshot $\mathcal{T}_{k}$ through the encoder and projection head. 

CLDG is trained to minimize local or global temporal translation invariance. The local and global contrastive learning loss of CLDG is as follows: 
\begin{equation}
\mathcal{L}^{\textrm{node} }=\sum_{i=1}^{N}\sum_{q=1}^{V}\sum_{k\neq q}^{V} \mathcal{L}_{i}^{\textrm{node} },
\label{eq:loss_node}
\end{equation}

\begin{equation}
\mathcal{L}^{\textrm{graph} }=\sum_{i=1}^{N}\sum_{q=1}^{V}\sum_{k\neq q}^{V} \mathcal{L}_{i}^{\textrm{graph} },
\label{eq:loss_graph}
\end{equation}
where $V$ is the multiple views sampled by the timespan view sampling layer. 

To sum up, in each training epoch, we first sample multiple different views through the timespan view sampling layer. Subsequently, $N$ nodes are sampled from the nodes shared by multiple views as a minibatch. Afterward, the representation of nodes and node neighborhoods in the minibatch are learned by using the encoder and projection head. Finally, the parameters in the encoder and projection head are updated by minimizing Eq.~\ref{eq:loss_node} or Eq.~\ref{eq:loss_graph}. 

\subsection{Advantages over Dynamic Graph Methods}
Compared with previous dynamic graph work, CLDG has the following advantages:

\textbf{Better Generality}. Existing dynamic graph work is generally designed solely for discrete-time dynamic graphs or continuous-time dynamic graphs. However, the timespan view sampling layer enables CLDG to handle both discrete-time and continuous-time dynamic graphs.

\textbf{Better Scalability}. The encoder module of CLDG can use any network architecture, and any effective encoder in the future can also be integrated into CLDG in a way similar to hot swap. 

\textbf{Lower Space and Time Complexity}. A generalized dynamic graph learning paradigm is to use graph neural networks to model structural information, and then use sequential models to explicitly model evolutionary information. For example, graph neural networks choose GCN, GAT, and sequence models choose RNN, LSTM, etc. However, CLDG implicitly exploits the additional temporal cues provided by dynamic graphs through contrastive learning, omitting the sequential model architecture. Therefore it has lower space and time complexity.

\begin{table}[htbp]\small
\renewcommand\arraystretch{1.2}
\setlength\tabcolsep{6pt}
\centering
  \caption{Statistics of the Datasets.}
  \label{tab:SD}
  \begin{tabular}{cccccccc}
    \toprule
    \# Dataset & \# nodes & \# temporal edges & \# classes \\
    \midrule
    DBLP & 25,387 & 185,480 & 10 \\
    Bitcoinotc & 5,881 & 35,592 & 3 \\
    TAX & 27,097 & 315,478 & 12 \\
    BITotc & 4,863 & 28,473 & 7 \\
    BITalpha & 3,219 & 19,364 & 7 \\
    TAX51 & 132,524 & 467,279 & 51 \\
    Reddit & 898,194 & 2,575,464 & 3 \\
    \bottomrule
  \end{tabular}
\end{table}

\section{Experiments}
In this section, we first introduce seven real-world dynamic graph datasets and twelve baselines. Then, we conduct extensive experiments to prove the CLDG yields consistent and significant improvements over baselines. Finally, ablation experiments on different components such as the timespan view sampling layer, base encoder layer, readout function, and contrastive loss function. The code and data are publicly available \footnote{\url{https://github.com/yimingxu24/CLDG}}.

The experiment aims to answer the following questions:

$\bullet$
$\mathbf{RQ_{1}}$: Does CLDG yield state-of-the-art results for unsupervised dynamic graph representation learning on established baselines?

$\bullet$
$\mathbf{RQ_{2}}$: How to choose the appropriate timespan as a contrastive view?

$\bullet$
$\mathbf{RQ_{3}}$: Does CLDG allow various encoder architecture choices without any limitations? Do different encoders have any effect on CLDG?

$\bullet$
$\mathbf{RQ_{4}}$: How about the time complexity and space complexity of CLDG?

$\bullet$
$\mathbf{RQ_{5}}$: How do various hyperparameters impact CLDG performance? 

\begin{table*}[]\small
\centering
\renewcommand\arraystretch{1.3} 
\setlength\tabcolsep{3pt}
\caption{Experimental results (\%) of classification tasks on seven datasets. We report both mean Accuracy and Weighted-F1, the input column highlights the data required for model training ($\mathbf{X}$ is the node features, $\mathbf{A}$ is the adjacency matrix, $\mathbf{S}$ is the diffusion matrix, $\mathbf{T}$ is the time information, and $\mathbf{Y}$ is the node labels). Bold represents the optimal in the unsupervised method, and underlined represents the global optimal.}
\begin{tabular}{ccccccccccccccccc}
\hline \hline
                              & \multirow{2}{*}{Method} & \multirow{2}{*}{Input} & \multicolumn{2}{c}{DBLP} & \multicolumn{2}{c}{Bitcoinotc} & \multicolumn{2}{c}{TAX} & \multicolumn{2}{c}{BITotc} & \multicolumn{2}{c}{BITalpha} & \multicolumn{2}{c}{TAX51} & \multicolumn{2}{c}{Reddit} \\
                              \cline{4-17}
                              &                         &                        & Acc     & Wei     & Acc        & Wei        & Acc     & Wei    & Acc      & Wei      & Acc       & Wei & Acc       & Wei & Acc       & Wei       \\ \hline
\multirow{4}{*}{\rotatebox{90}{Supervised}}   & LP                      & $\mathbf{A,Y}$                   & 52.18     & 51.54        & 41.97        & 31.25           & 35.67     & 30.88       & 60.24     & 50.28         & 76.97       & 67.18     & 28.82       & 24.36     & 66.71       & 60.88          \\
                              & GCN                     & $\mathbf{X, A, Y}$                & 71.35     & 71.08        & 54.61        & 54.41           & $\underline{71.65}$     & $\underline{71.37}$        & 59.24      & 50.52         & 76.19       & 73.65 & 40.42       & 33.64      & 67.83         & 63.26          \\
                              & GAT                     & $\mathbf{X, A, Y}$                & 70.01     & 69.21        & 52.46        & 50.60           & 62.00      & 59.78       & 62.59      & 48.34         & 80.21       & 71.40         & 38.82       & 30.21         & 69.31       & 58.01          \\
                              & GraphSAGE               & $\mathbf{X, A, Y}$                & $\underline{72.36}$     & $\underline{71.99}$        & 57.29        & 56.30           & 64.36     & 63.73       & 62.68      & $\underline{56.99}$         & 79.89       & $\underline{74.24}$         & $\underline{40.80}$       & $\underline{33.79}$     & 71.37       & \underline{65.74}          \\ \hline
\multirow{8}{*}{\rotatebox{90}{Unsupervised}} & CAW                     & $\mathbf{X, A, T}$                & 55.64     & 51.38        & $\underline{\mathbf{59.85}}$        & 57.92           & 53.25     & 47.11       & 63.56      & 53.59         & 77.64       & 71.99         & 36.52       & 29.52         & 67.79       & 57.80           \\
                              & TGAT                    & $\mathbf{X, A, T}$                & 57.48     & 51.96        & 58.56        & 55.73           & 50.12     & 45.62       & 62.53      & 49.53         & 77.63       & 68.03         & 34.50       & 28.44         & 65.57       & 56.81          \\
                              & DySAT                   & $\mathbf{X, A, T}$                & 54.12     & 52.48        & 51.32        & 48.80            & 52.31     & 51.42       & 62.01      & 48.70          & 77.61       & 68.07          & 23.81       & 22.89           & -     & -          \\
                              & MNCI                    & $\mathbf{X, A, T}$                & 67.03     & 66.46        & 55.14        & 54.79           & 45.44     & 37.13       & 63.49      & 54.66         & 79.04       & 71.80         & 39.13       & 32.29         & 70.94       & $\mathbf{65.26}$           \\
                              & DGI                 & $\mathbf{X, A}$                & 69.97     & 69.40        & 56.67        & 55.34           & 66.57     & 65.87       & 61.68      & 51.65         & 78.56       & $\mathbf{73.19}$     & 39.20       & 31.61     & 70.58       & 60.95          \\
                              & GRACE                 & $\mathbf{X, A}$                & 71.27     & 70.71        & 54.50        & 53.48           & 67.43     & 66.79       & 62.49      & 48.06         & 77.60       & 67.81           & -     & -           & -     & -          \\
                              & MVGRL                   & $\mathbf{X, A, S}$                & 71.32     & 70.91        & 55.31        & 54.54            & 67.70     & 67.08       & 59.62      & 53.00          & 75.26       & 72.71           & -     & -           & -     & -          \\
                              & CCA-SSG                 & $\mathbf{X, A}$                & 68.28     & 67.60        & 56.48        & 54.83           & 67.62     & 67.14       & 63.47      & 51.70         & 79.97       & 72.69         & 37.04       & 29.69         & 69.46       & 58.69           \\ 
                              & CLDG${_{node}}$                    & $\mathbf{X, A, T}$                & 71.80     & 71.55        & 59.17        & $\underline{\mathbf{58.45}}$           & $\mathbf{69.62}$     & $\mathbf{69.18}$       & $\underline{\mathbf{65.68}}$      & $\mathbf{54.74}$         & $\underline{\mathbf{80.61}}$       & 72.90         & $\mathbf{40.44}$       & $\mathbf{32.36}$       & $\underline{\mathbf{71.69}}$      & 62.87 \\          
                              & CLDG${_{graph}}$                 & $\mathbf{X, A, T}$                & $\mathbf{72.03}$     & $\mathbf{71.69}$        & 58.95        & 58.10           & 69.20     & 68.79       & 65.03      & 53.32         & 80.34       & 72.28         & 39.59       & 32.02         & 71.64       & 62.26  
                              \\ \hline
                              \hline
\end{tabular}
\label{tab:nc}
\end{table*}

\subsection{Experiment Preparation}
\subsubsection{Real-world Dataset}\label{dataset}
Seven datasets from four fields (i.e. academic citation network, tax transaction network, Bitcoin network, and reddit hyperlink network) are used to evaluate the quality of representations learned by CLDG. 
The statistics of the seven datasets are shown in Table~\ref{tab:SD}.

\textbf{Academic Citation Network}. The DBLP dataset is extracted from the DBLP \footnote{\url{http://dblp.uni-trier.de}}. The label is the research field of researchers. The label information divides researchers into ten different research areas, and each researcher has only one label. 

\textbf{TAX Transaction Networks}. The TAX and TAX51 company industry classification datasets are extracted from the tax transaction network data of two cities provided by the tax bureau. It consists of companies as vertices, and transaction relationships between companies as edges. The industry code of each company is provided by the tax bureau as labels of the dataset.

\textbf{Bitcoin Networks}. Bitcoin is a cryptocurrency used for anonymous transactions on the web. Due to the risk of trading with anonymity, this has led to the emergence of several exchanges where Bitcoin users rate how much they trust other users. Bitcoinotc, BITotc~\cite{kumar2016edge, liu2021inductive} and BITalpha~\cite{kumar2016edge, liu2021inductive} are three datasets from two bitcoin trading platforms, OTC and alpha~\footnote{\url{https://www.bitcoin-otc.com/} and \url{https://btc-alpha.com/}}. The label is the credit rating of the Bitcoin user.

\textbf{Reddit Hyperlink Network}. 
We construct the Reddit dataset from the Subreddit Hyperlink Network~\footnote{\url{http://snap.stanford.edu/data/soc-RedditHyperlinks.html}}. It consists of posts and hyperlinks in the community as nodes. The source and link relationship between each hyperlink and the post as edges. Each hyperlink is annotated with the sentiment of the source community post towards the target community post. This dataset can be used for sentiment classification.

\subsubsection{Baselines}
\ 
\newline
To verify that CLDG yields consistent and significant improvements, we compare CLDG with twelve state-of-the-art semi-supervised and unsupervised algorithms. Semi-supervised learning methods include label propagation algorithm and graph neural network algorithms:

\textbf{Semi-Supervised Methods:}

\noindent$\bullet$
\textbf{LP}~\cite{zhu2002learning}: LP iteratively adjusts the class in the unlabeled sample so that the class conditional distribution obtained allows a maximum a posteriori classification with minimum classification error on the labeled patterns.

\noindent$\bullet$
\textbf{GCN}~\cite{kipf2016semi}: GCN encodes graph structure and node features via a localized first-order approximation of spectral graph convolutions.

\noindent$\bullet$
\textbf{GAT}~\cite{velivckovic2017graph}: GAT uses self-attention to perform convolution operations in the spatial domain of the graph to learn node representations.

\noindent$\bullet$
\textbf{GraphSAGE}~\cite{hamilton2017inductive}: GraphSAGE samples the neighbors of each node in the graph network, then uses the aggregation function to obtain neighbor information, and finally learns the node representation.

We also compare with other unsupervised learning methods, including methods designed for dynamic graphs and contrastive learning methods:

\textbf{Unsupervised Dynamic Graph Methods:} 

\noindent$\bullet$
\textbf{CAW}\cite{wang2021inductive}: CAW implicitly extracts network motifs via temporal random walks and adopts set-based anonymization to establish the correlation between network motifs.

\noindent$\bullet$
\textbf{TGAT}\cite{xu2020inductive}: TGAT adapts the self-attention mechanism to handle the continuous time by proposing a time encoding, capturing temporal-feature signals in terms of both node and topological features on temporal graphs. 

\noindent$\bullet$
\textbf{DySAT}\cite{sankar2020dysat}: DySAT computes dynamic node representations through joint self-attention over the structural neighborhood and historical representations, thus capturing the temporal evolution of graph structure. 

\noindent$\bullet$
\textbf{MNCI}~\cite{liu2021inductive}: MNCI proposes an aggregator function that integrates neighborhood influence with community influence to generate node embeddings at any time. 

\textbf{Unsupervised Contrastive Learning Methods:} 

\noindent$\bullet$
\textbf{DGI}~\cite{velickovic2019deep}: DGI uses graph convolution network architecture to learn patch representations and corresponding high-level graph representations, and then maximizes the mutual information between them to train the model.

\noindent$\bullet$
\textbf{GRACE}~\cite{zhu2020deep}: GRACE generates two graph views by corrupting the structural and attribute levels, and learns node representations by maximizing the consistency of node representations in these two views.

\noindent$\bullet$
\textbf{MVGRL}~\cite{hassani2020contrastive}: MVGRL generates another view through graph diffusion and regular view input into two GNNs, and shared MLP to learn node representation, and learn graph representation through graph pooling layer. The discriminator scores the agreement between representations as a training signal.

\noindent$\bullet$
\textbf{CCA-SSG}~\cite{zhang2021canonical}: CCA-SSG first generates two views of the input graph through data augmentation, and then uses the idea based on CCA to maximize the correlation between the two views and encourages different feature dimensions to capture distinct semantics.

\subsubsection{Evaluation Protocols}
We divide the seven datasets according to the 1:1:8 split method of training set: validation set: test set.
Semi-supervised methods train the model through training and validation sets, outputting predicted labels for each node in the test set. 
Our method and other unsupervised methods follow a linear evaluation scheme as introduced in~\cite{velickovic2019deep}, where each model is firstly trained in an unsupervised manner. After training, freeze the model parameters and output the learned representations for all nodes. Subsequently, the representation is fed into a linear classifier, noting that only the node embeddings in the training set are used to train the classifier. We report the Accuracy and Weighted-F1 metrics for node classification in the test set.

\subsubsection{Implementation Details}
For CLDG, we implement the model with pytorch. Adam optimizer~\cite{kingma2014adam} is used in the training model stage and the training linear classifier stage. In the training model stage, the learning rate of the encoder and projection head is $4 \times 10^{-3}$, and the weight decay is $5 \times 10^{-4} $. In the training linear classifier stage, the learning rate is $10^{-2} $, the weight decay is $ 10^{-4} $. The base encoder adopts a two-layer GCN, the batch size is 256, the hidden layer dimension is 128, and the output dimension is 64.

\newcolumntype{x}[1]{>{\centering\arraybackslash}p{#1pt}}
\begin{table*}[h!]\small
	\label{tab:gs}
	\renewcommand\arraystretch{1.24}
	\caption{Comparison of different timespan view sampling architectures, sampling strategy, sampling view timespan size $s$ and number of sampling views $v$ in the DBLP and TAX dataset. The sampling strategy with low overlap rate between views, such as sequential and random, and sampling more views are beneficial to CLDG, and the timespan of sampling views is robust to CLDG.}
	
	\subfloat[ Dataset: \textbf{DBLP}, base encoder: \textbf{GCN}, epoch: \textbf{200}, loss function: \textbf{Eq.~\ref{eq:loss_node}}.
	\label{tab:gs_dblp}]{
    \begin{tabular}{x{14}|x{26}x{26}x{26}|x{26}x{26}x{26}|x{26}x{26}x{26}|x{26}x{26}x{26}}
		&  \multicolumn{3}{c|}{ \textbf{Sequential} } &  \multicolumn{3}{c|}{ \textbf{High Overlap Rate} }  &  \multicolumn{3}{c|}{ \textbf{Low Overlap Rate}} &  \multicolumn{3}{c}{ \textbf{Random}} \\
		\multicolumn{1}{c|}{\diagbox{v}{s}}  & 6 & 8 & 10 & 6 & 8 & 10 & 6 & 8 & 10 & 6 & 8 & 10  \\
		\shline
        2 & 71.11    & 71.01    & 70.84    & 69.44       & 68.92      & 69.16      & 70.26      & 70.59      & 70.10      & 71.01   & 70.69   & 70.48  \\
        3 & 71.34    & 71.10    & 71.17    & 69.92       & 69.78      & 69.56      & 70.85      & 70.73      & 70.79      & 70.95   & 71.06   & 70.75  \\
        4 & 71.40    & 71.34    & 71.00    & 70.01       & 69.75      & 69.82      & 70.80      & 70.60      & 70.73      & 70.53   & 71.06   & 70.61  \\
        5 & 71.55    & 71.49    & 71.23    & 70.48       & 70.62      & 70.23      & 71.16      & 70.96      & 70.54      & 71.04   & 70.77   & 70.74 \\
	\end{tabular}}
	\hfill
	
	\subfloat[ Dataset: \textbf{TAX}, base encoder: \textbf{GCN}, epoch: \textbf{200}, loss function: \textbf{Eq.~\ref{eq:loss_node}}.
	\label{tab:gs_tax}]{
    \begin{tabular}{x{14}|x{26}x{26}x{26}|x{26}x{26}x{26}|x{26}x{26}x{26}|x{26}x{26}x{26}}
		&  \multicolumn{3}{c|}{ \textbf{Sequential} } &  \multicolumn{3}{c|}{ \textbf{High Overlap Rate} }  &  \multicolumn{3}{c|}{ \textbf{Low Overlap Rate}} &  \multicolumn{3}{c}{ \textbf{Random}} \\
		\multicolumn{1}{c|}{\diagbox{v}{s}}  & 6 & 8 & 10 & 6 & 8 & 10 & 6 & 8 & 10 & 6 & 8 & 10  \\
		\shline
        2 & 67.05    & 67.94    & 65.75    & 63.52       & 62.87      & 62.68      & 64.98      & 64.22      & 63.51      & 65.26   & 65.13   & 65.08  \\
        3 & 67.65    & 67.25    & 66.58    & 63.79       & 63.68      & 63.04      & 64.53      & 64.76      & 63.91      & 66.33   & 65.57   & 65.63  \\
        4 & 68.99    & 67.85    & 67.51    & 64.22       & 63.69      & 63.40      & 64.91      & 64.53      & 64.05      & 66.80   & 66.46   & 65.97  \\
        5 & 68.98    & 67.01    & 67.60    & 64.26       & 63.35      & 63.50      & 65.03      & 64.76      & 64.11      & 66.79   & 66.18   & 66.19
	\end{tabular}}
	\hfill 
\end{table*}

\subsection{Comparison with State-of-the-Art ($\mathbf{RQ_{1}}$)}
To answer $\mathbf{RQ_{1}}$, we compare CLDG with 12 state-of-the-art algorithms on the seven dynamic graph datasets. 
The semi-supervised model in the baseline includes label propagation (LP), GCN, GAT, and GraphSAGE. We also compare with unsupervised dynamic graph models including CAW, TGAT, DySAT, and MNCI, and unsupervised contrastive learning methods including DGI, GRACE, MVGRL, and CCA-SSG. We use the Accuracy and Weighted-F1 as the evaluation metrics, bold indicates that the method is optimal among unsupervised methods, and underlined indicates that it is optimal among all baselines. We report classification results on 
seven datasets in Table~\ref{tab:nc}. We implement local-level and global-level versions of CLDG according to Eq.~\ref{eq:loss_node} and Eq.~\ref{eq:loss_graph}, both of which outperform other unsupervised methods. CLDG${_{graph}}$ performs better on the DBLP dataset, and CLDG${_{node}}$ performs better on the other six datasets. Among the Accuracy and Weighted-F1 metrics of the seven datasets, eleven metrics are optimal among the unsupervised methods. 
CLDG${_{node}}$ outperforms the previous state-of-the-art GraphSAGE model by 1.47\% on the average of all metrics.
In Bitcoinotc, BITotc, BITalpha and Reddit datasets, some metrics outperform all baselines. For example at Bitcoinotc, CLDG${_{node}}$ outperforms the best supervised method GraphSAGE by 2.15\% on Weighted-F1. On BITotc and BITalpha, CLDG${_{node}}$ outperforms the best supervised method GraphSAGE and GCN by 3.0\% and 0.4\% on Accuracy, respectively.

In summary, the experimental results proved that CLDG generalizes contrastive learning from static graphs to dynamic graphs through different levels of temporal translation invariance on graphs, and implicitly utilizes temporal information to achieve state-of-the-art results in unsupervised dynamic graphs representation learning.
    
\subsection{View Sampling Architecture ($\mathbf{RQ_{2}}$)}
The timespan view sampling module consists of three factors: sampling strategy, view timespan factor $s$ and the number of views sampled $v$. We design four different sampling strategies, sequential, high overlap rate, low overlap rate and random, and the details of the sampling strategy are demonstrated in this section~\ref{tpgs}. $v$ controls the number of the sampling timespan views, and $s$ controls the timespan size of the view. 
We show in Table~\ref{tab:gs_dblp} and Table~\ref{tab:gs_tax} how the three factors jointly affect the performance of CLDG on the DBLP and TAX datasets. 

The variation on the dynamic graph is continuous and smooth, and if the timespan views overlap physically, better results may be achieved by sharing more similar semantic contexts. Therefore, the main difference between the four sampling strategies is the overlap rate between views. However, the better performers of the four strategies are sequential and random. The high overlap rate strategy tends to be the worst performer, on average 1.41\% and 4.01\% lower than the sequential strategy in the DBLP and TAX datasets, respectively. This is a very interesting phenomenon, illustrating that on a dynamic graph with temporal translation invariance, a high overlap rate may lead to an overly simplistic contrastive learning task that makes the model non-robust. The sequential and randomized approaches maintain high-level semantic consistency rather than just low-level physical consistency allowing for better performance on the test set. The view timespan factor $s$, i.e., constructed views with different timespans, is relatively robust in both datasets and seems to have little effect on CLDG. In practice, we can choose some timespans with physical significance, such as day, week, month, and so on. Finally, on the number of timespan views v, we find that $v = 3,4,5$ on average 0.37\%, 0.34\%, 0.60\%, and 0.40\%, 0.87\%, 0.82\% higher than $v=2$, in DBLP and TAX datasets, respectively. This indicates that more timespan views are beneficial for CLDG, but the average improvement is smaller after the number of views exceeds 3.

In general, the sampling strategy with the low overlap rate between views and sampling more views are beneficial to CLDG, and the timespan of constructed views is robust.

\begin{table*}[]\small
\centering
\renewcommand\arraystretch{1.3}
\setlength\tabcolsep{3.61pt}
\caption{Comparison of space and time complexity of dynamic graph method on seven datasets. On the Reddit dataset, the parameters of CAW, TGAT, DySAT, and MNCI are 17,840.6, 17,813.8, 2.0 and 5,785.4 times of CLDG, respectively. On the TAX51 dataset, the running time of CLDG is 946.4, 158.2, 132.7 and 30.1 times faster than CAW, TGAT, DySAT, and MNCI. CLDG has lower space and time complexity.}
\begin{tabular}{ccccccccccccccccc}
\hline
\hline
  & \multirow{2}{*}{Method} & \multicolumn{2}{c}{DBLP} & \multicolumn{2}{c}{Bitcoinotc} & \multicolumn{2}{c}{TAX} & \multicolumn{2}{c}{BITotc} & \multicolumn{2}{c}{BITalpha} & \multicolumn{2}{c}{TAX51} & \multicolumn{2}{c}{Reddit}\\
  \cline{3-16}
  &                         & Param     & Time     & Param        & Time        & Param     & Time        & Param     & Time        & Param     & Time        & Param     & Time        & Param     & Time       \\ \hline
    & CAW       & 56.76M & 9403s & 13.39M & 3498s & 90.47M & 30702s & 11.31M & 2510s & 8.56M & 1768s & 156.32M & 329336s & 892.03M & 361580s \\
    & TGAT       & 55.41M & 5607s & 12.05M & 669s & 89.13M & 6067s & 9.96M & 1713s & 7.21M & 642s & 154.98M & 55066s & 890.69M & 50441s \\
    & DySAT & 0.10M & 5103s & 0.10M & 513s & 0.10M & 5781s & 0.10M & 422s & 0.10M & 456s & 0.10M & 46173s & 0.10M & - \\ 
    & MNCI & 8.22M & 4308s & 1.94M & 745s & 8.78M & 7580s & 1.62M & 550s & 1.09M & 481s & 42.72M & 10463s & 289.27M & 46586s \\ 
    & CLDG & \textbf{0.05M} & \textbf{297s} & \textbf{0.05M} & \textbf{136s} & \textbf{0.05M} & \textbf{312s} & \textbf{0.05M} & \textbf{74s} & \textbf{0.05M} & \textbf{248s}  & \textbf{0.05M} & \textbf{348s} & \textbf{0.05M} & \textbf{551s}\\     \hline
    \hline

\end{tabular}
\label{tab:STCA}
\end{table*}

\begin{figure*}
\centering
\subfloat[Effect of epoch]{
\label{fig:param1}
\includegraphics[width=4.2cm]{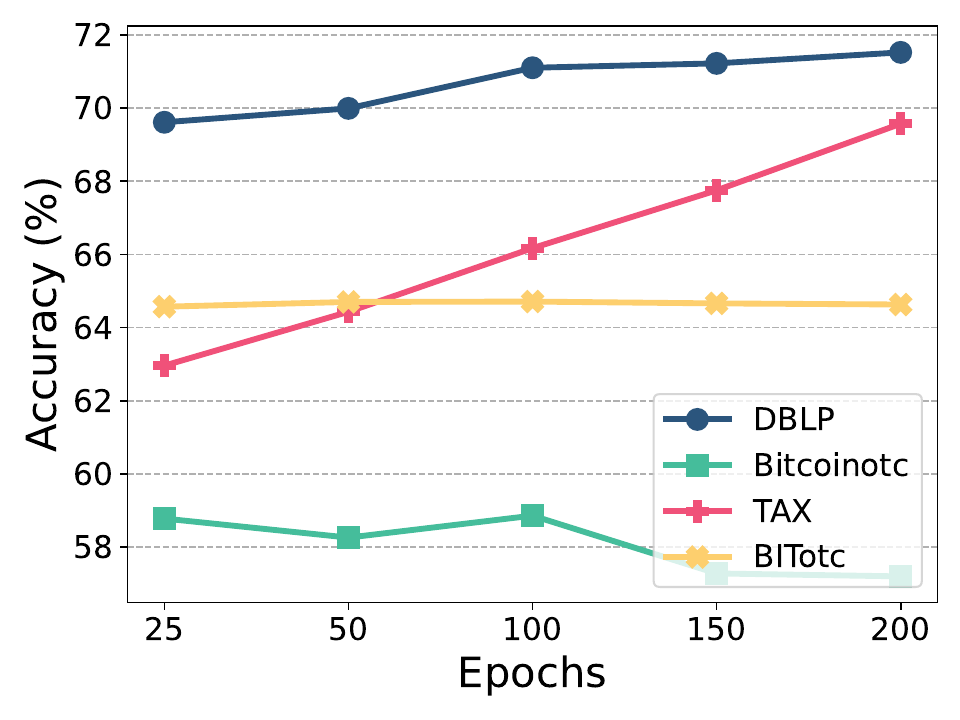}
}
\subfloat[Effect of batch size]{
\label{fig:param2}
\includegraphics[width=4.2cm]{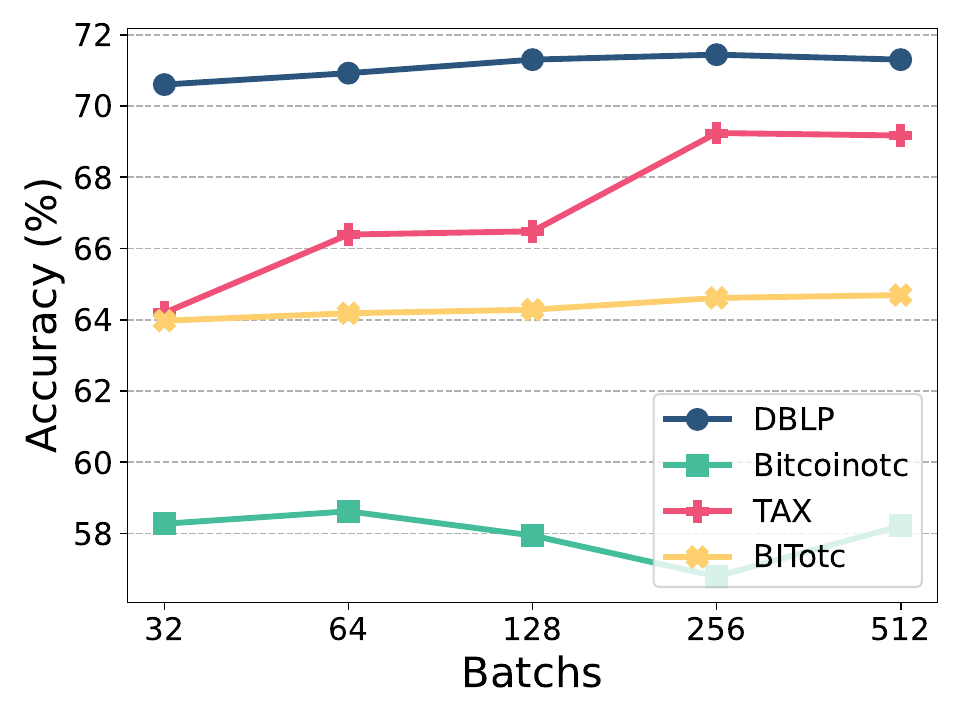}
}
\subfloat[Effect of dimension]{
\label{fig:param3}
\includegraphics[width=4.2cm]{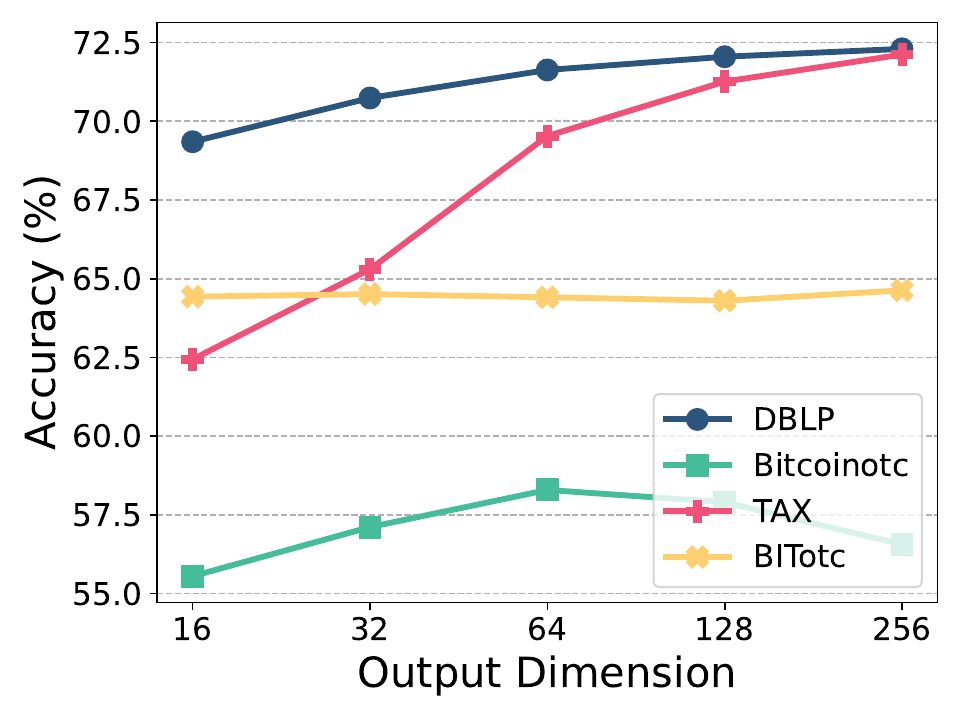}
}
\subfloat[Effect of layer]{
\label{fig:param4}
\includegraphics[width=4.2cm]{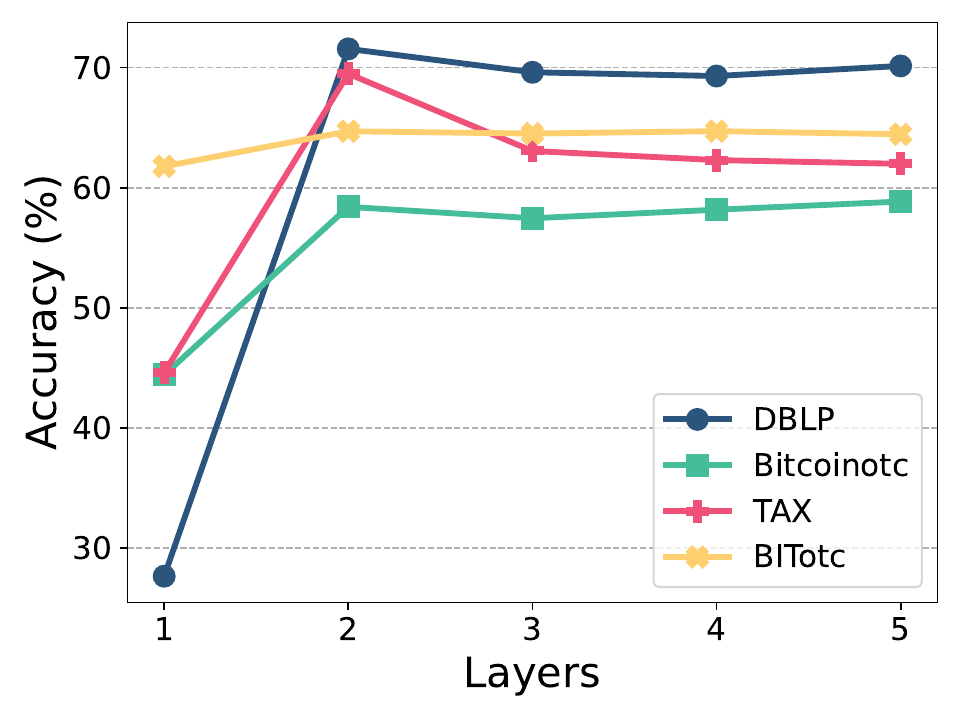}
}
\caption{Parameter sensitivity of CLDG. Effect of epoch, batch size, output dimension and layers on the node classification.}
\label{fig:parameters}
\end{figure*}

\begin{table}[]\small
\centering
\renewcommand\arraystretch{1.3}
\setlength\tabcolsep{3pt}
\caption{Comparison of different encoder architectures on three datasets. The number of layers in each encoder is 2, the timespan view sampling strategy is sequential, $s=4$, $v=2$, loss function is Eq.~\ref{eq:loss_node}, the epoch is 200, 100 and 100, respectively. CLDG allows a variety of encoder choices and is surprisingly robust. }
\begin{tabular}{ccccccccc}
\hline
\hline
  & \multirow{2}{*}{Encoder} & \multicolumn{2}{c}{DBLP} & \multicolumn{2}{c}{BITotc} & \multicolumn{2}{c}{BITalpha} \\
  \cline{3-8}
  &                         & Acc     & Wei-F1     & Acc        & Wei-F1        & Acc     & Wei-F1       \\ \hline
    & GCN       & 71.48 & 71.18 & 64.70 & 52.12 & 80.14 & 72.18 \\
    & GAT       & 73.49 & 73.16 & 64.94 & 52.57 & 80.12 & 71.79 \\
    & GraphSAGE & 73.19 & 72.51 & 64.06 & 50.75 & 80.09 & 71.97 \\ \hline
    \hline

\end{tabular}
\vspace{-4mm}
\label{tab:encoder}
\end{table}

\subsection{Encoder Architecture ($\mathbf{RQ_{3}}$)}
To answer $\mathbf{RQ_{3}}$, we implement three classical GNN encoders (GCN, GAT and GraphSAGE) for CLDG. We conduct experiments on the DBLP, BITotc and BITalpha datasets. The number of layers of the encoder is 2, the timespan view sampling strategy is sequential, $s$ is 4, $v$ is 2, the epoch is 200, 100 and 100, and the lambda is 0.5, 0.3 and 0.3, respectively. Table~\ref{tab:encoder} shows the classification results of different encoder architectures in the test set. The experimental results show that all three encoders have very competitive performance. Under the current parameter settings, GAT performs best on the DBLP and BITotc datasets, but only 0.30\% and 0.65\% higher than GraphSAGE in DBLP, and 0.24\% and 0.45\% higher than GCN in BITotc, on Accuracy and Weighted-F1, respectively. GCN is even only 0.02\% and 0.21\% higher than suboptimal in BITalpha, respectively.

To sum up, experiments demonstrate that CLDG allows various encoder architecture choices without limitations. All encoders show extremely competitive results, proving the surprising robustness of CLDG.

\subsection{Space and Time Complexity Analysis ($\mathbf{RQ_{4}}$)}
To answer $\mathbf{RQ_{4}}$, we compared CLDG with four other dynamic graph methods (CAW, TGAT, DySAT, and MNCI) in terms of space and time complexity. In Table~\ref{tab:STCA}, we measure the space and time complexity by the number of parameters and the model running time, respectively. In terms of space complexity, first, the parameters of CLDG do not increase with the increase of graph size. However, other dynamic graph models have a maximum of 104.21, 123.54, 1.00 and 265.39 times increase in parameters on the seven datasets, respectively. The CLDG parameters are derived from the base encoder and the projection head, and the base encoder can also be replaced with an encoder of lower space complexity and time complexity in the future. Then, the parameters of CLDG are much smaller than the existing SOTA dynamic graph model. The parameters of CAW, TGAT, DySAT, and MNCI are 17,840.6, 17,813.8, 2.0 and 5,785.4 times of CLDG on the Reddit dataset, respectively. Finally, in terms of time complexity, CLDG is implemented with a graph neighbor sampler, and the running time of CLDG is 946.4, 158.2, 132.7 and 30.1 times faster than CAW, TGAT, DySAT, and MNCI on TAX51 dataset.

In conclusion, CLDG has lower space and time complexity than existing dynamic graph methods, and is more significant on larger scale graphs. This makes CLDG easier to generalize to larger-scale graph learning.

\subsection{Hyperparameters Sensitivity ($\mathbf{RQ_{5}}$)}
We investigate the sensitivity of the parameters and report the Accuracy (\%) results for four datasets with various parameters in Figure~\ref{fig:parameters}.

$\bullet$
\textbf{Effect of epoch.} 
We first investigated the effect of epochs on CLDG. The results are shown in Fig.~\ref{fig:param1}, we can find that the performance of DBLP and TAX dataset keeps improving with the growth of epoch. Bitcoinotc and BITotc datasets have little difference in performance with increasing epochs. The reason is that larger datasets require more epochs to train, while smaller datasets only require fewer epochs to reach the optimal solution.

$\bullet$
\textbf{Effect of batch size.} 
We then investigated the effect of batch size on CLDG. The results are shown in Fig.~\ref{fig:param2}, on the four datasets, the classification performance generally improves with increasing batch size. On the Bitcoinotc dataset, when the batch size is 128 and 256, the performance decreases. The possible reason is that the samples of Bitcoinotc are unbalanced, which causes the nodes of the same label in a batch to be pulled apart in the Euclidean space. A more suitable temperature parameter $\tau$ might alleviate this problem.

$\bullet$
\textbf{Effect of output dimension.} 
Additionally, we investigate the effect of output dimension on CLDG. The results are shown in Fig.~\ref{fig:param3}, in the DBLP dataset and the TAX dataset, the classification performance improves with increasing dimensionality. In the Bitcoinotc dataset, the classification performance first increases and then decreases. The reason is that larger dimensions may introduce additional redundancy on smaller datasets.

$\bullet$
\textbf{Effect of model layers (the neighbors in how many hops are considered).} 
Finally, we study the effect of the number of model layers on CLDG, which aggregates $k$-hop neighbor information when Eq.~\ref{eq:encoder} stacks $k$ layers. The results are shown in Fig.~\ref{fig:param4}, when the number of layers is increased from 1 to 2, the classification performance is improved to varying degrees on each dataset, especially in the DBLP dataset. After increasing the number of layers of the model, the receptive field of the node increases, and the capability of the model first increases and then remains basically unchanged.

\section{Conclusions}
In this paper, we propose a simple and effective model CLDG, which generalizes contrastive learning to dynamic graphs. Specifically, CLDG first samples the input dynamic graph using a timespan view sampling layer. Subsequently, the local and global representations of each timespan view are learned through the encoder, projection head, and readout functions. Finally, the contrastive approach is used to maintain the temporal translation invariance of the local and the global. CLDG uses the additional temporal cues provided by dynamic graphs, thus avoiding the semantics and labels changes during the augmentation process. Experiments demonstrate the validity of the local and global temporal translation invariance assumptions. CLDG can achieve state-of-the-art unsupervised dynamic graph techniques and is competitive with supervised methods. Meanwhile, CLDG is very lightweight and scalable. Existing dynamic graph learning paradigms use graph neural networks and sequence models to model structural and evolutionary information, respectively. However, the main parameters and the training and inference time of CLDG depend on the base encoder. This allows CLDG to have lower space and time complexity, reducing the number of parameters and training time by up to 2,000.86 times and 130.31 times over existing models on the seven datasets. In addition, we also demonstrate the scalability of CLDG by replacing different base encoders. A limitation of our method is that it may not be applicable when the changes on the dynamic graph are non-continuous and non-smooth, and the labels on the graph are constantly changing.

\section*{Acknowledgment}

This research was partially supported by the National Key Research and Development Project of China No. 2021ZD0110700, the National Science Foundation of China under Grant Nos. 62002282, 62250009, 62037001, 61721002 and 62202029, "Pioneer" and "Leading Goose" R\&D Program of Zhejiang under Grant No. 2022C01107, the China Postdoctoral Science Foundation No. 2020M683492, the MOE Innovation Research Team No. IRT\_17R86, and Project of XJTU-SERVYOU Joint Tax-AI Lab.

\bibliographystyle{IEEEtran}
\bibliography{Reference}

\flushend
\end{document}